\pdfoutput=1

\documentclass[11pt]{article}

\usepackage[final]{acl}

\usepackage{times}
\usepackage{latexsym}

\usepackage[T1]{fontenc}

\usepackage[utf8]{inputenc}

\usepackage{microtype}

\usepackage{inconsolata}

\usepackage{amsfonts,amsmath,amssymb}
\usepackage{graphicx}
\usepackage{hyperref}
\usepackage{url}
\usepackage{multirow}
\usepackage{multicol}
\usepackage{cleveref}
\usepackage{paralist}
\usepackage{xspace}
\usepackage[normalem]{ulem}
\usepackage{booktabs}
\usepackage{rotating}
\usepackage{makecell}
\usepackage{paralist}
\usepackage{listings}
\usepackage{stfloats}
\usepackage{subcaption}
\usepackage{enumitem}
\usepackage{xcolor}
\usepackage{pifont}

\definecolor{codegreen}{rgb}{0,0.6,0}
\definecolor{codegray}{rgb}{0.5,0.5,0.5}
\definecolor{codepurple}{rgb}{0.58,0,0.82}
\definecolor{backcolour}{rgb}{0.95,0.95,0.92}
\definecolor{darkgreen}{rgb}{0.0, 0.7, 0.0}
\definecolor{darkred}{rgb}{0.7, 0.0, 0.0}

\newcommand{\cmark}{\textcolor{darkgreen}{\ding{51}}}%
\newcommand{\xmark}{\textcolor{red}{\ding{55}}}%

\lstdefinestyle{prompt_style}{
    frame=single,
    basicstyle=\ttfamily\scriptsize,
    backgroundcolor=\color{white},
    stringstyle=\color{black},
    commentstyle=\color{darkgreen}\slshape,
    stringstyle=\color{darkred},
    numberstyle=\tiny\color{codegray},
    emphstyle=\color{pink}\underbar,
    breakindent=0pt,
    escapeinside={(*@}{@*)},
    breakatwhitespace=true,
    breaklines=true,
    captionpos=b,
    keepspaces=true,
    numbersep=5pt,
    showspaces=false,                
    showstringspaces=false,
    showtabs=false,
    tabsize=2,
}

\lstdefinelanguage{json}{
    basicstyle=\ttfamily\scriptsize,
    stepnumber=1,
    numbersep=8pt,
    showstringspaces=false,
    breaklines=true,
    frame=single,
    keywordstyle=\bfseries\color{blue},
    stringstyle=\color{green!50!black},
    commentstyle=\color{gray},
    morestring=[b]",
    tabsize=2,
    literate=                   
     *{:}{{\textcolor{blue}{:}}}{1}
      {,}{{\textcolor{blue}{,}}}{1}
      {[}{{\textcolor{red}{[}}}{1}
      {]}{{\textcolor{red}{]}}}{1}
      {\{}{{\textcolor{red}{\{}}}{1}
      {\}}{{\textcolor{red}{\}}}}{1}
}

\crefname{equation}{Eq.}{Eq.}
\crefname{section}{Sec.}{Sec.}
\newcommand{\mscript}[1]{\text{\scriptsize{#1}}}

\newcommand{\ie}{\emph{i.e.,}\xspace}

\newcommand{\eg}{\emph{e.g.,}\xspace}

\newcommand{\benchname}{{AntiLeakBench}\xspace}

\title{
    AntiLeakBench: Preventing Data Contamination by Automatically Constructing Benchmarks with Updated Real-World Knowledge
}

\author{
    \textbf{Xiaobao Wu}$^{1,2}$\thanks{Work done during visiting at UCSB.} \quad
    \textbf{Liangming Pan}$^5$ \quad
    \textbf{Yuxi Xie}$^{3,2}$\footnotemark[1] \quad
    \textbf{Ruiwen Zhou}$^{4,2}$\footnotemark[1] \quad
    \textbf{Shuai Zhao}$^1$ \\
    \textbf{Yubo Ma}$^1$ \quad
    \textbf{Mingzhe Du}$^{1,3}$ \quad
    \textbf{Rui Mao}$^1$ \quad
    \textbf{Anh Tuan Luu}$^1$ \quad
    \textbf{William Yang Wang}$^2$ \\
    $^1$Nanyang Technological University \quad $^2$University of California, Santa Barbara \\
    $^3$National University of Singapore \quad $^4$Shanghai Jiao Tong University \; $^5$ University of Arizona \\
    \texttt{xiaobao002@e.ntu.edu.sg} \quad
    \texttt{william@cs.ucsb.edu}
}

\begin{document}
\maketitle

\begin{abstract}
    Data contamination hinders fair LLM evaluation by introducing test data into newer models' training sets.
    Existing studies solve this challenge by updating benchmarks with newly collected data.
    However, they fail to guarantee contamination-free evaluation
    as the newly collected data may contain pre-existing knowledge,
    and their benchmark updates rely on intensive human labor.
    To address these issues, we in this paper propose AntiLeakBench, an automated anti-leakage benchmarking framework.
    Instead of simply using newly collected data,
    we construct samples with explicitly new knowledge absent from LLMs' training sets,
    which thus ensures strictly contamination-free evaluation.
    We further design a fully automated workflow to build and update our benchmark without human labor.
    This significantly reduces the cost of benchmark maintenance to accommodate emerging LLMs.
    Through extensive experiments, we highlight that data contamination likely exists before LLMs' cutoff time
    and demonstrate that AntiLeakBench effectively overcomes this challenge.~\footnote{Our code and data are available at \url{https://github.com/bobxwu/AntiLeakBench}.}
\end{abstract}

\section{Introduction}
    In recent years, Large Language Models (LLMs) have demonstrated prominent capabilities in multiple fields \cite{radford2019language,touvron2023llama}.
    To thoroughly assess these capabilities, various benchmarks have been developed, such as MMLU \cite{hendrycks2021mmlu} and GSM8K \cite{cobbe2021gsm8k}.
    They are typically static and publicly accessible, serving as standardized tools to evaluate LLMs' performance.
    Unfortunately, the static nature of these benchmarks presents a significant challenge: \textbf{Data Contamination},
    where their test data may end up in newer LLMs' training sets.
    This issue can inflate model performance and thus undermine the reliability and validity of these benchmarks \cite{golchin2023time,roberts2023cutoff,deng2024investigating,dong2024generalization,jiang2024investigating}.
    To avoid data contamination, recent studies dynamically update the benchmarks
    by collecting new data released after LLM's knowledge cutoff time
    \cite{kiela2021dynabench,potts2021dynasent,kasai2023realtime,jain2024livecodebench}.
    For instance, LiveBench \cite{white2024livebench} collects newly released questions from math exams and code platforms like LeetCode.

\begin{figure}[!t]
    \centering
    \includegraphics[width=\linewidth]{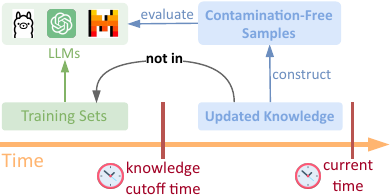}
    \caption{
        Illustration of \benchname.
        It constructs contamination-free samples
        based on the knowledge updated after LLMs' cutoff time,
        which thus are not in LLMs' training sets.
    }
    \label{fig_illustration}
\end{figure}

    However, despite their prevalence,
    we emphasize these benchmarks encounter two limitations:
    \begin{inparaenum}[(\bgroup\bfseries i\egroup)]
        \item
            \textbf{Weak guarantee for contamination-free evaluation.}
            They ignore to verify whether the newly collected data contain truly new knowledge \cite{roberts2023cutoff,white2024livebench,jain2024livecodebench}.
            For example, exam questions or coding problems from LeetCode may be later republished or referenced.
            As a result, the knowledge of these data may overlap with LLMs' training,
            which potentially leads to data contamination.
        \item
            \textbf{High dependency on human labor for maintenance.}
            Their benchmark updates often require intensive human labor, such as annotating the collected data \cite{gu2024xiezhi,xu2024let}.
            In consequence, this significantly hinders their frequent maintenance,
            especially in light of the rapid emergence of new LLMs.
            For instance, RealTimeQA \cite{kasai2023realtime} and KoLA \cite{yu2023kola} have barely been updated recently.
    \end{inparaenum}
    Together these limitations undermine the reliability and practicality of existing benchmarks for contamination-free evaluation.

    To address these limitations, we in this paper propose \textbf{\benchname}, an automated anti-leakage benchmarking framework to prevent data contamination.
    As illustrated in \Cref{fig_illustration},
    rather than directly collecting newly released data as previously,
    we identify new real-world knowledge updated after LLM's cutoff time.
    Then we construct question-answering samples querying these updated knowledge,
    accompanied by their corresponding real-world supporting documents.
    This ensures the updated knowledge is absent from LLMs' training sets,
    and thus the constructed samples on them are \textbf{strictly contamination-free}.

    Furthermore, we design \textbf{a fully automated workflow} to build and update \benchname.
    It eliminates the need for human labor, enabling the benchmark to be seamlessly updated to accommodate emerging LLMs.
    As such, this significantly reduces the maintenance cost of our benchmark, enhancing its practicality and scalability.

    We evaluate a series of LLMs on our \benchname with samples before and after the cutoff time.
    We observe that their performance commonly drops after the cutoff time.
    This trend highlights the likely data contamination in LLM evaluation.
    The experimental results additionally manifest the effectiveness of \benchname for contamination-free evaluation.
    The contributions of this paper can be concluded as follows:
    \begin{itemize}[leftmargin=*, itemsep=0pt]
        \item
            We propose \benchname, an automated anti-leakage benchmarking framework,
            ensuring contamination-free evaluation by constructing test samples with updated real-world knowledge.
        \item
            We propose an automated building workflow that automatically builds and updates \benchname without human labor,
            enabling to easily accommodate emerging new LLMs.
        \item
            We conduct extensive experiments involving various LLMs on multiple tasks
            and demonstrate the effectiveness of \benchname for contamination-free evaluation.
    \end{itemize}

\section{Related Work}
    \paragraph{Data Contamination}
        Many benchmarks have been widely used to assess the impressive capabilities of LLMs
        across various tasks, like question-answering, reading comprehension, and math reasoning
        \cite{pan2023fact,pan2024fallacy,zhou2024rulearena,wu2025sailing,zhao2025survey}.
        Notable examples include ARC \cite{clark2018think}, MMLU \cite{hendrycks2021mmlu}, BIG-bench \cite{srivastava2022bigbench}, and GSM8K \cite{cobbe2021gsm8k}.
        Although they have played a significant role,
        their static nature may cause the data contamination issue \cite{magar2022data,yang2023rethinking,golchin2023data,jacovi2023stop,li2023open,oren2023proving,sainz2023nlp,ni2024training}.
        Due to this,
        some work discloses several benchmarks are gradually less effective 
        \cite{schaeffer2023pretraining,zhou2023don}.
        For example, evidence shows some LLMs have overfitted to the GSM8K, compromising its validity~\cite{zhang2024careful}.

    \paragraph{Contamination-Free Evaluation}
        To achieve contamination-free evaluation, recent studies dynamically update benchmarks by collecting new data \cite{zhu2023dyval,thrush2022dynatask,qian2024varbench,srivastava2024functional}.
        For instance, RealTimeQA \cite{kasai2023realtime} periodically collects multi-choice quizzes from newspapers.
        Similarly, LiveCodeBench \cite{jain2024livecodebench} frequently crawls programming questions from code platforms like LeetCode.
        LiveBench \cite{white2024livebench} extends them
        by covering more domains, such as math competitions, research papers, and news articles.
        They cannot guarantee contamination-free evaluation since they rely solely on the newly released data
        and also need intensive human labor for construction \cite{liska2022streamingqa,mousavi2024dyknow}.
        Recent ADU \cite{ying2024automating} updates existing benchmarks by paraphrasing data through LLMs, but it may risk introducing mistakes and biases into evaluation.
        Different from these studies,
        our \benchname constructs samples with newly updated real-world knowledge to ensure contamination-free evaluation.
        It also introduces a fully automated building workflow without human labor.
        These differences make \benchname a more reliable and practical benchmarking framework for consistent contamination-free evaluation.

\begin{figure*}[!ht]
    \centering
    \includegraphics[width=\linewidth]{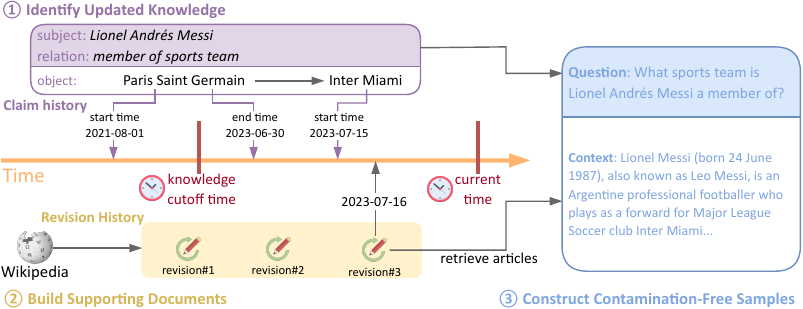}
    \caption{
        Illustration of the automated benchmark building workflow without human labor.
        After data preparation, it includes three main steps:
        (1) Identify updated knowledge after the cutoff time;
        (2) Build supporting documents;
        (3) Construct contamination-free samples (\Cref{fig_multihop} exemplifies how to construct multi-hop samples).
    }
    \label{fig_workflow}
\end{figure*}

\section{\benchname}
    In this section, we present how to automatically build \benchname with updated knowledge.
    \Cref{fig_workflow} illustrates the building workflow.
    We mainly focus on question-answering tasks regarding the updated knowledge for evaluation.
    This is because they allow precise control over the knowledge being evaluated, thus ensuring contamination-free evaluation.
    In contrast, using other tasks like summarization and code generation poses two challenges:
    \begin{inparaenum}[(\bgroup\bfseries i\egroup)]
        \item
            Identifying the specific knowledge embedded in them is inherently complex, making strict contamination-free unfeasible.
        \item
            Their intricate context structures make it difficult to automatically synthesize new high-quality samples without introducing errors and inconsistencies.
    \end{inparaenum}

    \subsection{Preparing Data}
        We begin by preparing the data to build the benchmarks.
        For the knowledge source, we leverage Wikidata \cite{wikidata2014}, a widely used knowledge base that is frequently updated by contributor communities.
        Wikidata provides an extensive repository of real-world factual claims involving numerous entities.
        Each claim is represented as a triplet \textit{(subject, relation, object)}, such as \textit{(Lionel Andrés Messi, member of sports team, Paris Saint Germain)}.

        We sample a subset of relations associated with physical entities, such as \textit{member of sports team};
        we exclude the less meaningful relations, for example, ones about virtual entities, like \textit{geometry coordinates}, similar to \citet{zhong2023mquake,wu2024akew} (See details in \Cref{app_workflow}).
        For each claim, we extract two qualifiers from Wikidata: \textit{start time} and \textit{end time},
        which specifies when the claim starts and ends, \eg the timeline of a football player in a team.
        We use these prepared data to build and update our \benchname.

    \subsection{Identifying Updated Knowledge}
        We then identify updated knowledge based on the prepared data.
        Considering $t_{1}$ as the knowledge cutoff time of an LLM and $t_{2}$ as the current time,
        our goal is to find the updated knowledge that occurs after $t_{1}$ and before $t_{2}$.
        For this purpose, we figure out the history of claims.
        Specifically, we group all the claims by their subject and relation
        and sort the claims in each group chronologically based on their start time.
        If a new claim emerges after the cutoff time $t_{1}$ in the history, \ie the object changes,
        we identify the new claim as the updated knowledge.
        For instance, in \Cref{fig_workflow} the object of (\textit{Lionel Andrés Messi}; \textit{member of sports team}) shifts after the cutoff time: \textit{Paris Saint Germain $\rightarrow$ Inter Miami}.
        From this shift, we extract updated knowledge with the new claim: \textit{(Lionel Andrés Messi; member of sports team; Inter Miami)}.
        We emphasize that the LLM is unaware of this knowledge because it occurs after its cutoff time.

        One may wonder what if the object changes twice and eventually reverts to its original one, like a player returning to his previous team.
        To exclude this case, we additionally confirm that the new claim is different from the old one
        and then consider it as updated knowledge.

    \subsection{Building Supporting Documents}
        To provide context for the updated knowledge, we build supporting documents from real-world sources.
        While LLMs could generate such documents, this may introduce mistakes or biases as LLMs often hallucinate or misinterpret information \cite{white2024livebench}.
        To maintain accuracy and reliability, we rely on the well-maintained and widely trusted Wikipedia as the source of supporting documents.
        This choice is further justified since the updates of Wikidata commonly follow the updates of Wikipedia \cite{wikidata2014}.

        In detail, we denote the identified updated knowledge as a claim $(s_{1}, r_{1}, o_{1})$
        and retrieve the Wikipedia page revision history of either its subject $s_{1}$ or object $o_{1}$, determined by its relation,
        \eg subject \textit{Lionel Andrés Messi} for relation \textit{member of sports team} in \Cref{fig_workflow}.
        Then we find the revision made after the start time of the updated knowledge.
        With this revision, we retrieve its corresponding article from Wikipedia
        and check if the summary of the article contains both the subject and object (or their aliases).
        If true, we consider this article as a supporting document for the updated knowledge (We show their validity in \Cref{sec_human_verifycation}).
        For example, \Cref{fig_workflow} illustrates a Wikipedia article indicating \textit{Lionel Andrés Messi} is
        a member of \textit{Inter Miami}.
        Here the supporting document is revised after LLMs' cutoff time, so it is also nonexistent in their training sets.

    \subsection{Constructing Contamination-Free Samples} \label{sec_construct_samples}
        Now we construct test samples querying the above updated knowledge $(s_{1}, r_{1}, o_{1})$,
        with its supporting document as context, denoted as $D$.
        Since the knowledge and the supporting document are both absent from LLMs' training sets,
        the constructed test samples for them are strictly contamination-free.
        This is different from previous studies: they overlook verifying if the inherent knowledge of their samples does not exist in LLM's training sets.

        \paragraph{Tasks}
            We mainly focus on question-answering tasks, including both single-hop and multi-hop questions.
            Following the common practice \cite{yang2018hotpotqa,kovcisky2018narrativeqa,bai2024longbench},
            each sample is denoted as $(Q, C, A)$ with question $Q$, context $C$, and expected answer $A$.
            \begin{itemize}[leftmargin=*, itemsep=2pt]
                \item
                    \textbf{Single-Hop}.
                    In this task, we directly ask what is the answer to the updated knowledge by providing the supporting document.
                    We first consider \textbf{Single-Hop Gold}:
                    Question $Q$ is formulated with predefined templates based on the relation in the updated knowledge,
                    Context $C$ only includes the supporting document $D$, and Answer $A$ is the object $o_{1}$ and its aliases.
                    For instance,
                    in \Cref{fig_workflow} the question is \textit{What sports team is Lionel Andrés Messi a member of?};
                    the context is \textit{Lionel Andrés Messi}'s Wikipedia article; the answer is the name and aliases of his sports team.

                    To enhance the difficulty, we further introduce a new task: \textbf{Single-Hop} $N_{\mscript{d}}$:
                    Question $Q$ remains the same, and we augment Context $C$ with $N_{\mscript{d}}$ distracting documents.
                    Here these distracting documents are randomly sampled from other samples' supporting documents which do not contain the subject $s_{1}$ and object $o_{1}$ of the updated knowledge.
                    As such, this task further assesses the long-context capability of LLMs,
                    specifically locating relevant information within distractors.

\begin{figure}[!t]
    \centering
    \includegraphics[width=\linewidth]{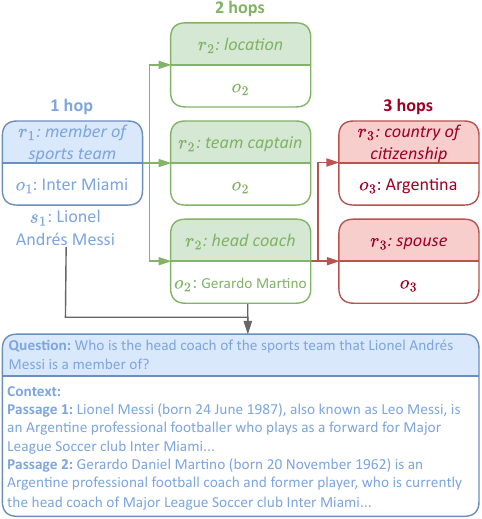}
    \caption{
        Illustration of constructing multi-hop samples
        by finding the consequent relations of previous objects.
    }
    \label{fig_multihop}
\end{figure}

                \item
                    \textbf{Multi-Hop}.
                    Moreover, we construct multi-hop questions for harder reasoning.
                    We first design \textbf{Multi-Hop Gold}.
                    Specifically, starting with the updated knowledge $(s_{1}, r_{1}, o_{1})$,
                    we build a chain of $H$ connecting claims:
                    $((s_{1}, r_{1}, o_{1}), (s_{2}, r_{2}, o_{2}), \dots, (s_{H}, r_{H}, o_{H}))$.
                    Here the object of the $i$-th triplet serves as the subject of the $(i\!+\!1)$-th triplet, \ie $o_{i} = s_{i+1}$.
                    As such, Question $Q$ is formulated across this chain,
                    Context $C$ includes the supporting document of each knowledge in this chain,
                    and Answer $A$ is the last object $o_{H}$.
                    For instance, \Cref{fig_workflow} shows a multi-hop question \textit{Who is the coach of the sports team that Lionel Andrés Messi is a member of?};
                    the context is \textit{Lionel Andrés Messi}'s and \textit{Gerardo Martino}'s Wikipedia articles;
                    the answer is the last object \textit{Gerardo Martino}.

                    Similarly, we also consider \textbf{Multi-Hop} $N_{\mscript{d}}$:
                    the context additionally covers $N_{\mscript{d}}$ randomly sampled supporting documents of other samples as the distracting documents.
                    This task assesses the multi-hop reasoning ability of LLMs, requiring them to connect and integrate information across a long context with distractors.
        \end{itemize}

        \paragraph{Question Formats}
            We consider two common question formats.
            \begin{itemize}[leftmargin=*, itemsep=2pt]
                \item
                    \textbf{Generation.}
                    Question $Q$ solely is the question as aforementioned.
                \item
                    \textbf{Multi-Choice.}
                    In this format, we additionally prompt LLMs with 4 options and ask them to select one.
                    We design the 4 options as follows:
                    \begin{inparaenum}[(\bgroup\bfseries a\egroup)]
                        \item
                            \textbf{Correct option}, \ie the correct answer to the question.
                        \item
                            \textbf{Unknown option}, represented as a string ``Unknown''.
                        \item
                            \textbf{Outdated option}. The old answer before the cutoff time, \eg \textit{Paris Saint Germain}) in \Cref{fig_workflow}.
                        \item
                            \textbf{Noise option}.
                            A randomly sampled, unrelated answer from other samples.
                    \end{inparaenum}
                    For multi-hop questions, we ignore finding outdated options and instead provide two noise options to accelerate the building workflow.
            \end{itemize}

        The above introduces the automated workflow to build our \benchname.
        \Cref{tab_example} shows an example of Single-Hop Gold, and more examples are in \Cref{app_examples}.

\begin{table*}[!t]
    \centering
    \setlength{\tabcolsep}{6mm}
    \renewcommand{\arraystretch}{1.2}
    \resizebox{\linewidth}{!}{
        \begin{tabular}{lcccl}
        \toprule
        \textbf{Benchmark} & \textbf{Strictly Contamination-Free} & \textbf{Automated} & \textbf{Multilingual} & \textbf{Data Source} \\
        \midrule
        Realtime QA\cite{kasai2023realtime} & \xmark& \xmark& \xmark& Real world \\
        LiveBench\cite{white2024livebench} & \xmark& \xmark& \xmark& Real world \\
        ADU\cite{ying2024automating} & \xmark& \cmark & \xmark& LLM generation \\
        \midrule
        \textbf{\benchname} & \cmark & \cmark & \cmark & Real world \\
        \bottomrule
        \end{tabular}%
    }
    \caption{
        Comparisons between \benchname and other benchmarking frameworks.
    }
    \label{tab_comparison}
\end{table*}

    \subsection{Benchmark Maintenance}
        Our \benchname supports easy maintenance.
        We only need to download the latest Wikidata dump and then execute our automated workflow to update benchmarks.
        The whole process requires no human labor, and hence we can effortlessly maintain the benchmarks for newer LLMs.

    \subsection{Multilingual Benchmarks}
        Moreover, \benchname features multilingual evaluation.
        It can seamlessly produce samples in various languages
        via our automated workflow with the multilingual nature of both Wikidata and Wikipedia.
        This enables us to evaluate LLMs in various linguistic contexts.
        See more details in \Cref{app_workflow}.

\begin{table}[!t]
    \centering
    \setlength{\tabcolsep}{3mm}
    \renewcommand{\arraystretch}{1.2}
    \resizebox{\linewidth}{!}{
        \begin{tabular}{p{6em}p{15em}}
        \toprule
        \textbf{Attributes} & \multicolumn{1}{l}{\textbf{Examples}} \\
        \midrule
        {question\newline{}(generation)} & What sports team is Lionel Andrés Messi a member of? \\
        \midrule
        {answer\newline{}(generation)} & Inter Miami CF\newline{}Inter Miami\newline{}Club Internacional de Fútbol Miami \\
        \midrule
        {question\newline{}(multi-choice)} & What sports team is Lionel Andrés Messi a member of?\newline{}A. Inter Miami CF\newline{}B. Paris Saint-Germain F.C.\newline{}C. Prime Minister of Romania\newline{}D. Unknown. \\
        \midrule
        {answer\newline{}(multi-choice)} & A \\
        \midrule
        {subject} & Lionel Messi\newline{}Lionel Andres Messi\newline{}Lionel Andrés Messi \\
        \midrule
        pid   & P54 (member of sports team) \\
        \midrule
        {object} & Inter Miami CF\newline{}Inter Miami\newline{}Club Internacional de Fútbol Miami \\
        \midrule
        {object\_old} & Paris Saint-Germain F.C.\newline{}Paris Saint-Germain Football Club\newline{}Paris Saint-Germain FC \\
        \midrule
        context & Lionel Andrés Messi (; born 24 June 1987), also known as Leo Messi, is an Argentine professional footballer who plays as a forward for Major League Soccer club Inter Miami… \\
        \bottomrule
        \end{tabular}%
}
    \caption{
        An example from \benchname.
    }
    \label{tab_example}
\end{table}

\begin{table}[!t]
    \centering
    \renewcommand{\arraystretch}{1.2}
    \resizebox{\linewidth}{!}{
        \begin{tabular}{lrr}
        \toprule
        \textbf{Quality Metrics} & \textbf{Single-Hop Gold} & \textbf{Multi-Hop Gold} \\
        \midrule
        Context Accuracy & 97.3  & 98.7 \\
        Answer Accuracy & 96.7  & 97.3 \\
        \bottomrule
        \end{tabular}%
    }
    \caption{
        Data quality by human verification.
    }
    \label{tab_data_quality}
\end{table}

    \subsection{Data Quality Verification} \label{sec_human_verifycation}
        To verify the data quality of produced samples in \benchname, we conduct human verifications.
        We ask human annotators to evaluate the accuracy of both answers and contexts in the samples.
        See \Cref{app_data_verification} for the verification procedure and agreement analysis.
        \Cref{tab_data_quality} shows that \benchname achieves a high standard of data quality with question and context accuracy over 96\%.

\begin{figure*}[!ht]
    \centering
    \begin{subfigure}[c]{0.5\linewidth}
        \centering
        \includegraphics[width=\linewidth]{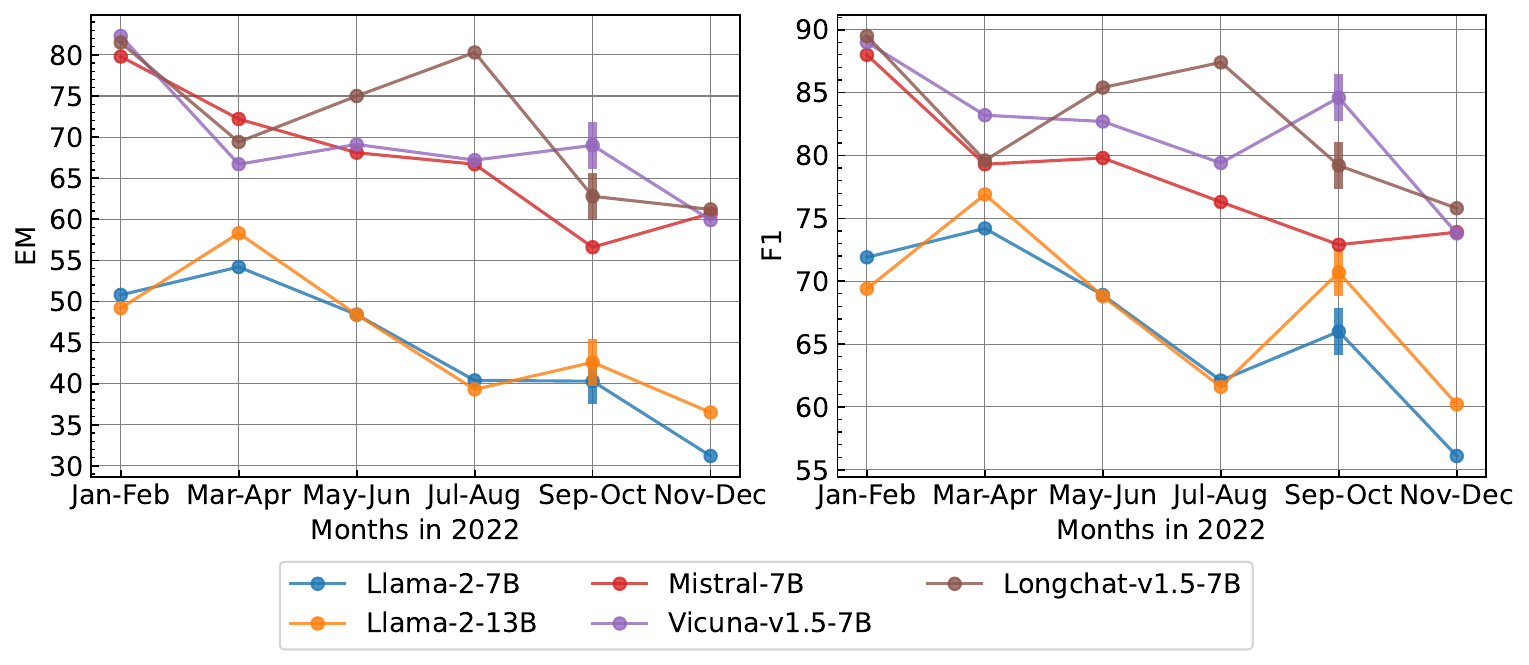}
    \end{subfigure}%
    \begin{subfigure}[c]{0.5\linewidth}
        \centering
        \includegraphics[width=\linewidth]{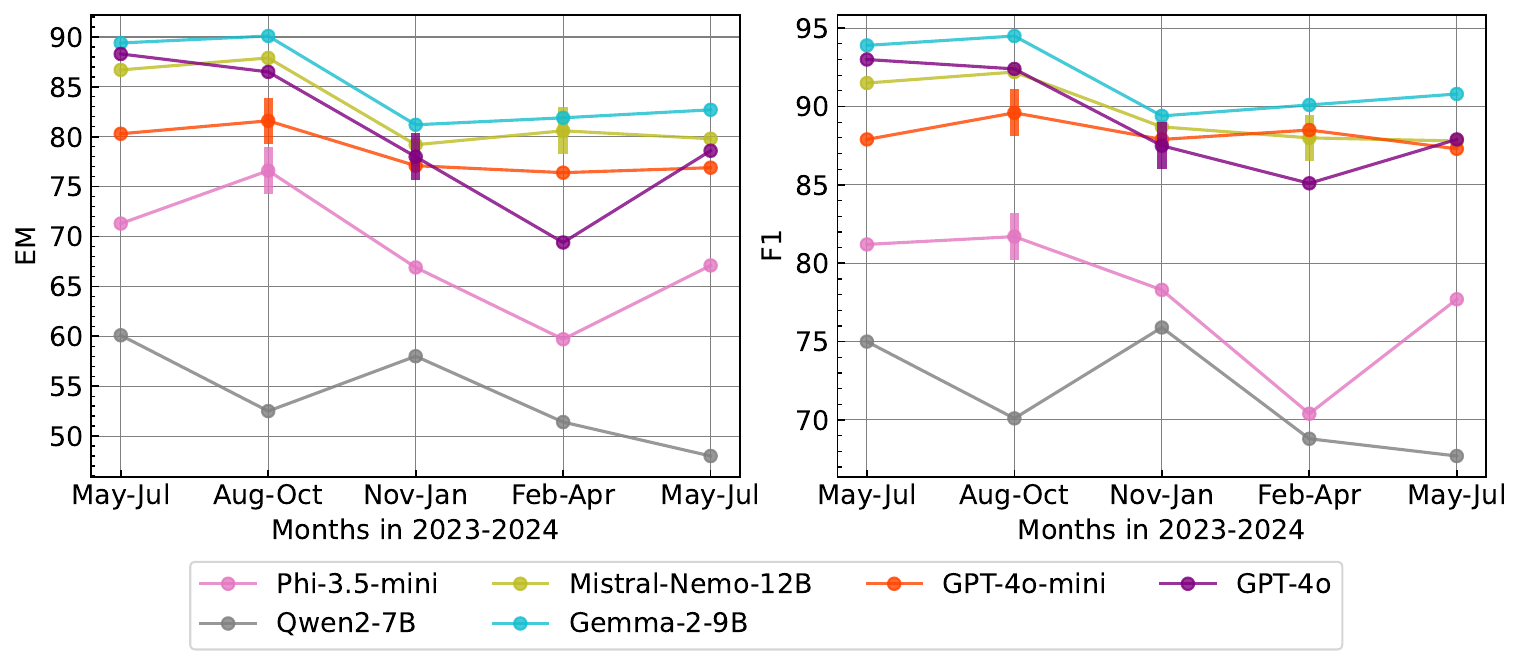}
    \end{subfigure}
    \caption{
        EM and F1 performance at each time interval.
        Marker $\;\raisebox{-0.3mm}{\rule{0.5mm}{3.3mm}}\;$ denotes LLM's cutoff time.
    }
    \label{fig_contamination_gen}
\end{figure*}

\begin{figure*}[!ht]
    \centering
    \begin{subfigure}[c]{0.5\linewidth}
        \centering
        \includegraphics[width=\linewidth]{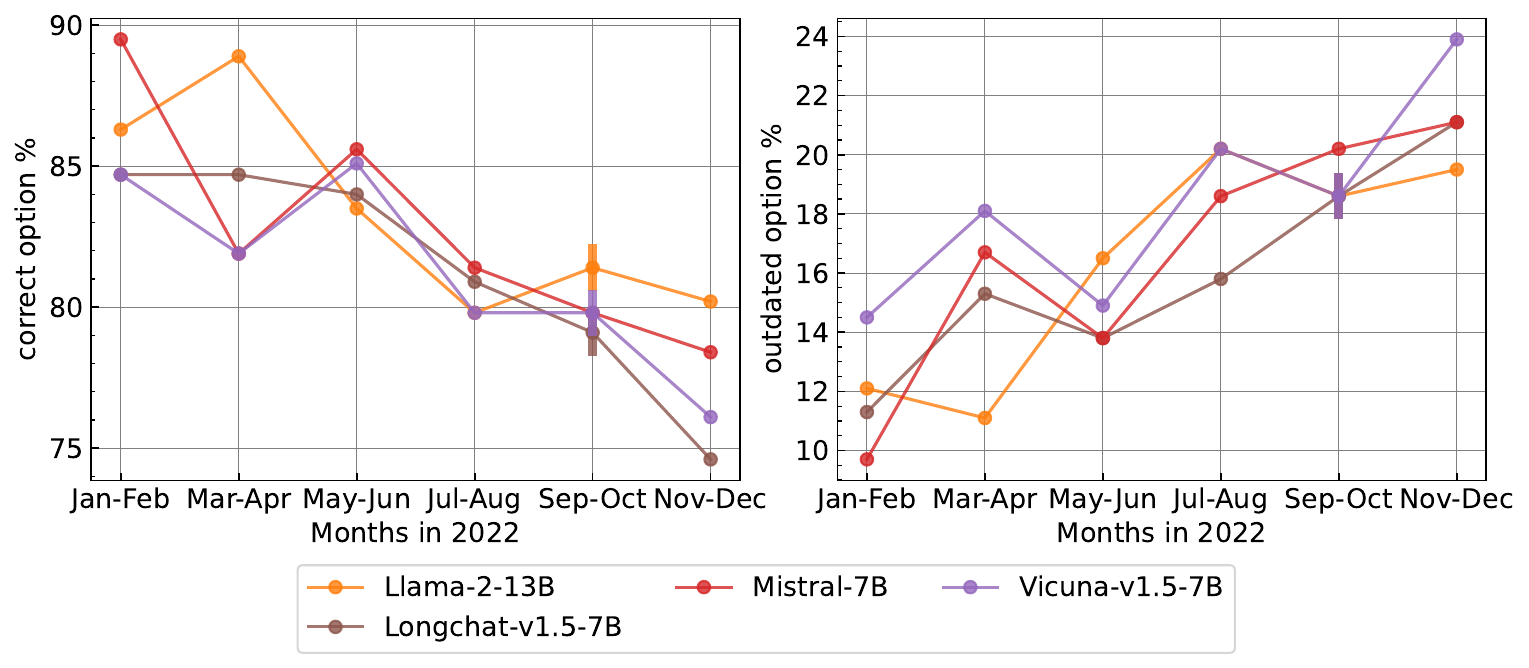}
    \end{subfigure}%
    \begin{subfigure}[c]{0.5\linewidth}
        \centering
        \includegraphics[width=\linewidth]{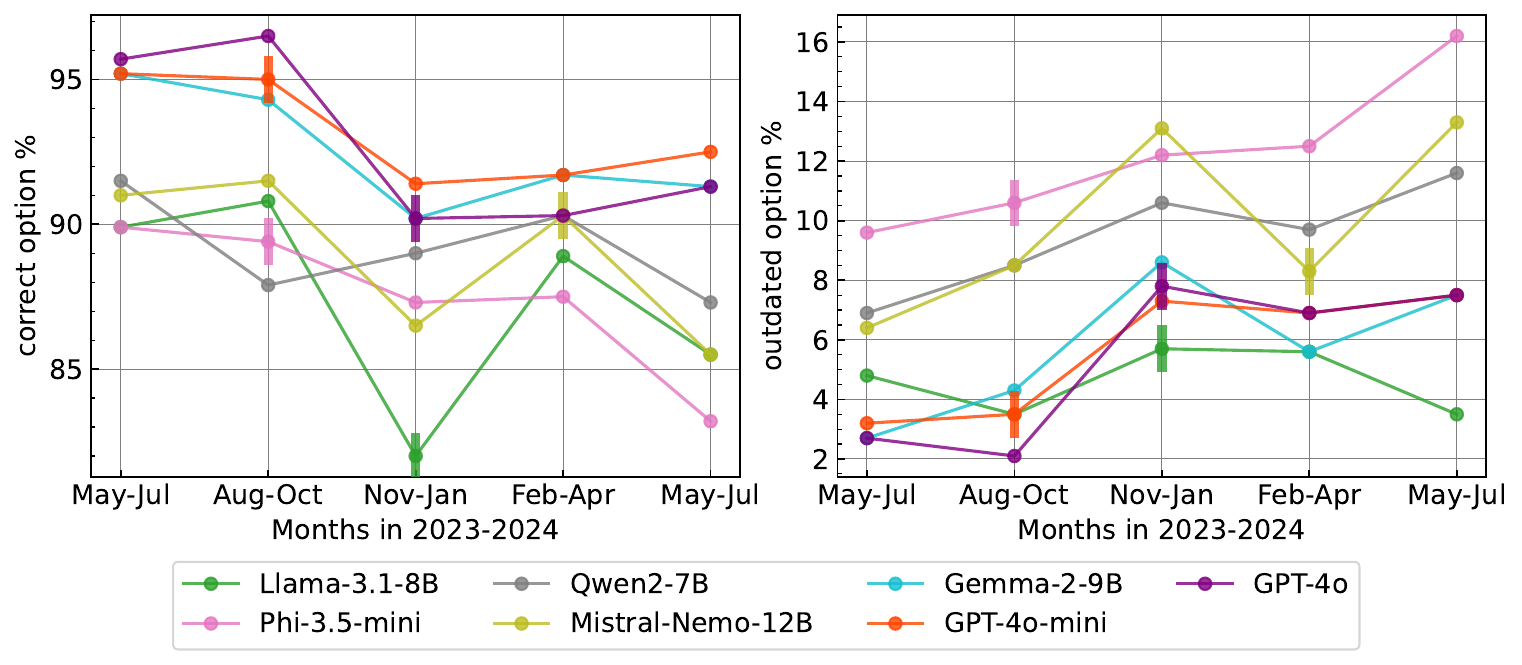}
    \end{subfigure}
    \caption{
        Correct and outdated option proportions at each time interval.
        Marker $\;\raisebox{-0.3mm}{\rule{0.5mm}{3.3mm}}\;$ denotes LLM's cutoff time.
    }
    \label{fig_contamination_multichoice}
\end{figure*}

    \subsection{Comparisons with Existing Benchmarks}
        \Cref{tab_comparison} compares our \benchname with other benchmarks.
        We highlight its vital advantages:
        \begin{inparaenum}[(\bgroup\bfseries i\egroup)]
            \item
                \textbf{Strictly contamination-free}.
                \benchname ensures that the constructed samples must cover updated knowledge absent from LLMs' training sets,
                which guarantees strictly contamination-free evaluation.
            \item
                \textbf{Automated workflow}.
                Our fully automated building workflow avoids human labor,
                which significantly reduces the cost of maintaining the benchmark for newer LLMs.
                In consequence, this fortifies its adaptivity and long-term applicability.
            \item \textbf{Multilingual}.
                Our method supports the construction of multilingual samples.
                This enables us to extensively assess LLMs' capabilities across different languages.
            \item
                \textbf{Real-world data}.
                We build test samples from real-world data sources like Wikipedia, rather than LLM-generated content.
                This grounds the benchmark in practical and authentic data.
        \end{inparaenum}
        With these advantages, \benchname serves as a reliable testbed for evaluating LLMs in a strictly contamination-free environment.

\section{Experiment}
    In this section, we experiment with different LLMs on our \benchname to  show its effectiveness and investigate data contamination.

    \subsection{Experiment Setup}
        \paragraph{Large Language Models}
            We experiment with the following 12 common language models:
            Llama-2-7B \cite{touvron2023llama},
            Llama-2-13B \cite{touvron2023llama},
            Mistral-7B \cite{jiang2023mistral},
            Vicuna-v1.5-7B \cite{vicuna2023},
            LongChat-v1.5-7B \cite{longchat2023},
            Llama-3.1-8B \cite{dubey2024llama},
            Phi-3.5-mini \cite{abdin2024phi},
            Qwen-2-7B \cite{yang2024qwen2},
            Mistral-Nemo-12B \cite{jiang2023mistral},
            Gemma-2-9B \cite{gemma_2024}.
            We also consider proprietary models: GPT-4o-mini and GPT-4o \cite{achiam2023gpt4o}.
            \Cref{tab_LLM_date} summarizes their release and knowledge cutoff time.
            We use the prompts in \cref{app_prompt} following \citet{bai2024longbench},
            which \textbf{explicitly ask LLMs to use the provided context to answer questions.}

\begin{table*}[!ht]
    \centering
    \renewcommand{\arraystretch}{1.2}
        \resizebox{\linewidth}{!}{
        \begin{tabular}{l|rr|rr|rr|rr|rr|rr|rr|rr|rr}
        \toprule
        \multirow{3}[6]{*}{\textbf{\makecell[l]{Language\\Models}}} & \multicolumn{8}{c|}{\textbf{Single-Hop}}                      & \multicolumn{8}{c|}{\textbf{Multi-Hop}}                       & \multicolumn{2}{c}{\multirow{2}[4]{*}{\textbf{Avg}}} \\
        \cmidrule{2-17}      & \multicolumn{2}{c|}{Gold} & \multicolumn{2}{c|}{$N_{\mscript{d}}$=3} & \multicolumn{2}{c|}{$N_{\mscript{d}}$=5} & \multicolumn{2}{c|}{$N_{\mscript{d}}$=7} & \multicolumn{2}{c|}{Gold} & \multicolumn{2}{c|}{$N_{\mscript{d}}$=3} & \multicolumn{2}{c|}{$N_{\mscript{d}}$=5} & \multicolumn{2}{c|}{$N_{\mscript{d}}$=7} & \multicolumn{2}{c}{} \\
        \cmidrule{2-19}      & \textbf{EM} & \textbf{F1} & \textbf{EM} & \textbf{F1} & \textbf{EM} & \textbf{F1} & \textbf{EM} & \textbf{F1} & \textbf{EM} & \textbf{F1} & \textbf{EM} & \textbf{F1} & \textbf{EM} & \textbf{F1} & \textbf{EM} & \textbf{F1} & \multicolumn{1}{c}{\textbf{EM}} & \multicolumn{1}{c}{\textbf{F1}} \\
        \midrule
        Llama-2-7B & 40.6  & 63.5  & 16.8  & 41.2  & 11.6  & 30.9  & 9.4   & 24.5  & 33.6  & 50.2  & \textbf{19.4} & \textbf{32.2} & \textbf{15.8} & \textbf{28.1} & \textbf{12.2} & \textbf{22.7} & 19.9  & 36.7 \\
        Llama-2-13B & 42.7  & 65.3  & 14.0  & 40.6  & 9.4   & 30.6  & 7.0   & 24.0  & 13.3  & 34.6  & 4.1   & 21.5  & 2.7   & 17.8  & 2.3   & 15.2  & 11.9  & 31.2 \\
        Mistral-7B & 65.4  & 77.2  & 27.8  & 41.3  & 16.7  & 27.3  & 7.3   & 15.3  & 21.4  & 27.9  & 11.5  & 17.2  & 8.1   & 14.3  & 6.5   & 11.1  & 20.6  & 29.0 \\
        Vicuna-v1.5-7B & 66.8  & 79.9  & 39.1  & 60.4  & 25.8  & 48.3  & 15.3  & 39.1  & 26.0  & 43.5  & 11.1  & 22.9  & 8.1   & 19.5  & 5.4   & 15.7  & 24.7  & 41.2 \\
        Longchat-v1.5-7B & \textbf{75.5} & \textbf{84.5} & \textbf{58.2} & \textbf{72.8} & \textbf{47.6} & \textbf{65.5} & \textbf{37.0} & \textbf{56.3} & \textbf{38.8} & \textbf{51.4} & 17.6  & 30.6  & 12.0  & 25.8  & 4.7   & 3.9   & \textbf{36.4} & \textbf{48.9} \\
        \midrule
        Llama-3.1-8B & 19.2  & 66.2  & 21.4  & 59.4  & 18.1  & 53.5  & 14.2  & 45.7  & 24.4  & 50.2  & 11.7  & 33.0  & 9.4   & 27.5  & 6.8   & 21.9  & 15.6  & 44.7 \\
        Phi-3.5-mini & 69.0  & 78.7  & 34.0  & 40.5  & 26.5  & 33.7  & 15.2  & 22.2  & 45.4  & 59.7  & 20.8  & 29.5  & 14.9  & 21.1  & 9.8   & 14.4  & 29.4  & 37.5 \\
        Qwen-2-7B & 54.8  & 72.4  & 15.5  & 38.5  & 9.8   & 26.6  & 7.2   & 21.2  & 35.9  & 48.3  & 23.7  & 33.4  & 18.1  & 26.1  & 13.6  & 20.1  & 22.3  & 35.8 \\
        Mistral-Nemo-12B & 82.7  & 89.7  & 75.6  & 83.8  & 66.3  & 75.1  & 51.8  & 62.2  & 57.7  & 67.3  & 39.1  & 47.7  & 33.8  & 41.4  & 24.0  & 29.0  & 53.9  & 62.0 \\
        Gemma-2-9B & \textbf{85.0} & \textbf{91.6} & 80.2  & 86.2  & 68.8  & 75.2  & 55.4  & 61.2  & \textbf{82.7} & \textbf{86.4} & 63.0  & 68.3  & 55.8  & 61.2  & 49.0  & 53.5  & 67.5  & 73.0 \\
        GPT-4o-mini & 78.5  & 88.1  & 80.3  & 89.2  & 79.1  & 88.1  & 79.2  & 88.5  & 68.8  & 83.1  & 60.5  & 75.3  & 57.1  & 73.1  & 54.2  & 70.6  & 69.7  & 82.0 \\
        GPT-4o & 81.2  & 89.5  & \textbf{84.1} & \textbf{90.8} & \textbf{83.5} & \textbf{90.3} & \textbf{84.8} & \textbf{91.4} & 71.5  & 85.9  & \textbf{71.9} & \textbf{86.1} & \textbf{70.2} & \textbf{84.8} & \textbf{70.2} & \textbf{84.8} & \textbf{77.2} & \textbf{87.9} \\
        \bottomrule
        \end{tabular}%
    }
    \caption{
        EM (Exact Match) and F1 results in the \textbf{generation} format on \benchname.
        Gold means only gold documents; $N_{d}$ is the number of distracting documents.
        The best is in \textbf{bold}.
    }
    \label{tab_main_gen}
\end{table*}

        \paragraph{Constructing Test Samples}
            To investigate the impact of data contamination,
            we on purpose construct two kinds of test samples:
            \begin{inparaenum}[(\bgroup\bfseries i\egroup)]
                \item
                    \textbf{Pre-cutoff samples}, containing knowledge before the cutoff time.
                    These samples may already exist in the LLMs' training sets.
                \item
                    \textbf{Post-cutoff samples}, containing knowledge updated after the cutoff time.
                    These samples are absent from LLMs' training sets.
            \end{inparaenum}

            To construct these samples, we establish two time periods according to the LLMs' cutoff time summarized in \Cref{tab_LLM_date}:
            \begin{inparaenum}[(\bgroup\bfseries i\egroup)]
                \item
                    From 2022-01-01 to 2023-01-01, divided into 2-month intervals.
                    This targets the first 5 models since their knowledge cutoff time mostly falls around Sep 2022,
                    such as Vicuna-v1.5-7B.
                \item
                    From 2023-05-01 to 2024-08-01, divided into 3-month intervals.
                    Similarly, this is tailored for the last 7 models
                    whose cutoff time mostly falls around the end of 2023,
                    \eg Llama-3.1-8B and GPT-4o.
            \end{inparaenum}
            Here we divide these periods into shorter intervals
            because (i) this can ensure each interval contains a sufficient number of samples, enabling statistically meaningful analysis,
            and (ii) this provides detailed granular insights into how contamination risks and model performance evolve over time.            
            We report the statistics of the produced samples of our benchmark in \Cref{tab_stats_size,tab_stats_word_count}.
            See more building details in \Cref{app_workflow}.

        \paragraph{Evaluation Metrics}
            For the generation format (\Cref{sec_construct_samples}), we use \textbf{EM} (Extract Matching) and token-based \textbf{F1} scores, following the standard practice in question answering \cite{rajpurkar2016squad}.
            For the multi-choice format, we use \textbf{Acc} (Accuracy) and \textbf{F1} scores following \citet{kasai2023realtime}.

\begin{table*}[!ht]
    \centering
    \renewcommand{\arraystretch}{1.2}
    \resizebox{\linewidth}{!}{
        \begin{tabular}{l|rr|rr|rr|rr|rr|rr|rr|rr|rr}
        \toprule
        \multirow{3}[6]{*}{\textbf{\makecell[l]{Language\\Models}}} & \multicolumn{8}{c|}{\textbf{Single-Hop}}                      & \multicolumn{8}{c|}{\textbf{Multi-Hop}}                       & \multicolumn{2}{c}{\multirow{2}[4]{*}{\textbf{Avg}}} \\
        \cmidrule{2-17}      & \multicolumn{2}{c|}{Gold} & \multicolumn{2}{c|}{$N_{\mscript{d}}$=3} & \multicolumn{2}{c|}{$N_{\mscript{d}}$=5} & \multicolumn{2}{c|}{$N_{\mscript{d}}$=7} & \multicolumn{2}{c|}{Gold} & \multicolumn{2}{c|}{$N_{\mscript{d}}$=3} & \multicolumn{2}{c|}{$N_{\mscript{d}}$=5} & \multicolumn{2}{c|}{$N_{\mscript{d}}$=7} & \multicolumn{2}{c}{} \\
        \cmidrule{2-19}      & \textbf{Acc} & \textbf{F1} & \textbf{Acc} & \textbf{F1} & \textbf{Acc} & \textbf{F1} & \textbf{Acc} & \textbf{F1} & \textbf{Acc} & \textbf{F1} & \textbf{Acc} & \textbf{F1} & \textbf{Acc} & \textbf{F1} & \textbf{Acc} & \textbf{F1} & \multicolumn{1}{c}{\textbf{Acc}} & \multicolumn{1}{c}{\textbf{F1}} \\
        \midrule
        Llama-2-7B & 41.7  & 30.7  & 3.7   & 5.6   & 3.5   & 5.3   & 2.8   & 5.4   & 18.7  & 30.9  & 6.8   & 9.9   & 5.6   & 8.1   & 3.6   & 6.9   & 10.8  & 12.9 \\
        Llama-2-13B & \textbf{82.1} & \textbf{82.2} & 73.7  & 73.6  & 60.1  & 59.9  & 51.7  & 51.3  & \textbf{97.5} & \textbf{97.5} & \textbf{88.5} & \textbf{88.5} & \textbf{82.8} & \textbf{83.1} & 75.2  & 75.2  & 76.5  & 76.4 \\
        Mistral-7B & 81.8  & 81.8  & 65.9  & 65.8  & 58.3  & 58.2  & 52.3  & 52.3  & 88.7  & 88.6  & 77.2  & 77.2  & 72.7  & 72.8  & 67.7  & 67.2  & 70.6  & 70.5 \\
        Vicuna-v1.5-7B & 80.1  & 80.0  & \textbf{75.6} & \textbf{75.4} & \textbf{73.1} & \textbf{72.9} & \textbf{69.6} & \textbf{69.4} & 96.8  & 96.9  & 84.0  & 84.2  & 82.6  & 83.0  & \textbf{77.0} & \textbf{77.2} & \textbf{79.8} & \textbf{79.9} \\
        Longchat-v1.5-7B & 79.6  & 79.7  & 68.5  & 68.8  & 65.1  & 51.8  & 62.3  & 61.2  & 93.2  & 93.4  & 76.7  & 78.0  & 70.4  & 71.5  & 66.6  & 68.0  & 72.8  & 71.6 \\
        \midrule
        Llama-3.1-8B & 86.7  & 90.4  & 62.2  & 74.0  & 48.9  & 62.9  & 37.8  & 52.9  & 70.5  & 81.4  & 50.7  & 64.8  & 40.9  & 56.2  & 30.8  & 44.9  & 53.6  & 65.9 \\
        Phi-3.5-mini & 87.4  & 87.5  & 85.6  & 85.8  & 84.7  & 85.4  & 79.6  & 82.5  & 96.5  & 97.0  & 85.3  & 86.2  & 78.0  & 80.3  & 68.6  & 72.3  & 83.2  & 84.6 \\
        Qwen-2-7B & 89.1  & 39.7  & 83.0  & 27.9  & 78.2  & 24.6  & 77.0  & 78.5  & 97.6  & 98.3  & 94.5  & 54.2  & 92.4  & 46.4  & 91.5  & 91.7  & 87.9  & 57.7 \\
        Mistral-Nemo-12B & 88.5  & 71.1  & 88.8  & 71.8  & 84.7  & 70.2  & 77.8  & 83.8  & 91.1  & 94.6  & 77.1  & 68.4  & 69.9  & 64.0  & 43.1  & 58.7  & 77.6  & 72.8 \\
        Gemma-2-9B & 92.4  & 92.4  & 86.7  & 86.5  & 76.9  & 61.6  & 69.4  & 69.3  & 97.1  & 97.1  & 88.3  & 88.3  & 81.8  & 65.4  & 77.4  & 77.4  & 83.8  & 79.8 \\
        GPT-4o-mini & \textbf{93.2} & \textbf{93.2} & \textbf{93.8} & \textbf{93.8} & 93.3  & 93.3  & 93.5  & 93.5  & \textbf{98.5} & \textbf{98.5} & \textbf{96.4} & \textbf{96.4} & \textbf{95.4} & \textbf{95.4} & 93.5  & 93.5  & \textbf{94.7} & \textbf{94.7} \\
        GPT-4o & 92.8  & 92.8  & 93.5  & 93.5  & \textbf{94.0} & \textbf{94.0} & \textbf{94.0} & \textbf{94.0} & 97.9  & 97.9  & 95.8  & 95.8  & \textbf{95.4} & \textbf{95.4} & \textbf{93.9} & \textbf{93.9} & 94.7  & 94.7 \\
        \bottomrule
        \end{tabular}%
    }
    \caption{
        Acc and F1  results in the \textbf{multi-choice} format on \benchname.
        Gold means only gold documents; $N_{d}$ is the number of distracting documents.
        The best is in \textbf{bold}.
    }
    \label{tab_main_multichoice}
\end{table*}

    \subsection{Data Contamination Analysis} \label{sec_data_contamination}
        We first analyze the impact of data contamination
        by looking into the performance trends of LLMs.
            \Cref{fig_contamination_gen} presents the trends of EM and F1 scores under the Single-Hop Gold task in the generation format
            (Llama-3.1-8B is excluded due to its low EM; See \Cref{sec_performance_analysis}).
            We observe a general performance decline for most LLMs after their knowledge cutoff time, 
            although all samples are constructed in the same way.
            For instance, Vicuna-v1.5-7B's EM and F1 start dropping near its cutoff time Sep 2022;
            Similarly, GPT-4o-mini remains stable until its cutoff time Oct 2023 and decreases thereafter.
            According to this decline, we conclude two key findings:
            \begin{enumerate}[wide, labelindent=0pt, labelsep=2pt, itemsep=1pt, topsep=1pt, label=\textbf{(\arabic*)}]
                \item
                    \textbf{
                    Pre-cutoff samples come with data contamination, which inflates LLMs' performance.
                    }
                    Pre-cutoff samples may overlap with LLMs' training sets.
                    This allows LLMs to correctly handle these samples based on their prior knowledge rather than actually understanding the provided context.
                    As a result, evaluating models solely on pre-cutoff samples can overestimate their capabilities.
                \item
                    \textbf{Contamination-free post-cutoff samples are more challenging and can more accurately assess LLMs.}
                    Post-cutoff samples are free from contamination since they include knowledge updated after LLMs' cutoff time.
                    To deal with these samples, LLMs must demonstrate true comprehension and reasoning over the given context and questions,
                    as they cannot rely on prior knowledge alone.
                    As such, evaluating with post-cutoff samples more accurately reflects LLMs' abilities.
            \end{enumerate}

            Interestingly, we notice that some LLMs experience performance drops even before the cutoff time.
            This is probably because their training data cover less knowledge close to the cutoff time.
            Knowledge sources such as news articles and Wikipedia pages usually become widely documented only after initial events occur.

            Besides, \Cref{fig_contamination_multichoice} plots the proportions of correct and outdated options selected under the Single-Hop Gold task in the multi-choice format
            (Llama-2-7B is excluded due to its low performance; See \Cref{sec_performance_analysis}).
            Notably LLMs increasingly favor outdated options over correct ones (See option types in \Cref{sec_construct_samples}).
            For example, Mistral-Nemo-7B's proportion of selecting correct options decreases from 91.0\% to 85.5\%, while the proportion of outdated options rises from 6.4\% to 13.3\%.
            These results again validate that pre-cutoff samples, due to data contamination, could inflate the performance.

        In summary, the above results demonstrate the effectiveness of our \benchname, which effectively ensures contamination-free evaluation
        and serves as a more reliable assessment of LLMs.

    \subsection{Overall Performance Analysis} \label{sec_performance_analysis}
        Next we analyze the overall performance of LLMs.
        \Cref{tab_main_gen,tab_main_multichoice} summarize the average results over all samples across tasks in the generation and multi-choice format, respectively.
        According to these results, we have the following observations.
        \begin{enumerate}[wide, labelindent=0pt, labelsep=2pt, itemsep=1pt, topsep=1pt, label=\textbf{(\arabic*)}]
            \item
                \textbf{\benchname poses a significant challenge for LLMs.}
                \Cref{tab_main_gen} shows that most models score EM and F1 below 50, indicating a substantial gap between their capabilities and the benchmark's requirements.
                Only two models, GPT-4o-mini and GPT-4o, reach EM and F1 scores around 70 and 80.
                These results highlight the challenging nature of our \benchname for evaluating LLMs.
            \item
                \textbf{Proprietary models lead in performance.}
                \Cref{tab_main_gen,tab_main_multichoice} reveal that proprietary models surpass open-source ones by a large margin.
                For instance, GPT-4o achieves the highest average EM and F1 scores, 77.2 and 87.9, respectively,
                while the runner-up open-source model only has 53.9 and 62.0.
                Additionally, GPT-4o's performance remains relatively stable despite the increasing number of distracting documents.
                This substantial margin may be attributed to their longer max context length, superior model architectures, and larger parameter sizes.
        \end{enumerate}

    \subsection{Difficulty Analysis} \label{sec_difficulty_analysis}
        We further analyze the difficulty concerning tasks and question formats based on \Cref{tab_main_gen,tab_main_multichoice}.
        We summarize the key findings as follows.
        \begin{enumerate}[wide, labelindent=0pt, labelsep=2pt, itemsep=1pt, topsep=1pt, label=\textbf{(\arabic*)}]
            \item
                \textbf{Multi-choice format is significantly easier.}
                LLMs generally perform better in the multi-choice format compared to the generation format.
                For example, GPT-4o-mini achieves Acc and F1 scores over 90, surpassing its performance in the generation format.
                This is because the multi-choice format simplifies the tasks by providing explicit answer options,
                enabling LLMs to identify correct answers with partial comprehension.
            \item
                \textbf{Longer contexts bring about higher difficulty.}
                LLMs' performance gradually decreases along with more distracting documents, such as from $N_{\mscript{d}}$=3 to 7 in \Cref{tab_main_gen}.
                The reason is that long contexts distract LLMs from locating relevant information.
                This underscores the necessity of LLMs' stronger ability to handle long contexts.
            \item
                \textbf{Multi-hop tasks are more challenging.}
                Compared to single-hop ones, LLMs' performance becomes lower on multi-hop tasks.
                For example, in \Cref{tab_main_gen} the EM score of Longchat-v1.5-7B decreases from 75.5 on the Single-Hop Gold to 38.8 on the Multi-Hop Gold;
                GPT-4o-mini decreases from 78.5 to 68.8.
                These multi-hop tasks require LLMs to decompose complex questions and reason across long and interconnected contexts.
        \end{enumerate}

\section{Conclusion}
    In this paper, we propose \benchname, an automated anti-leakage benchmarking framework.
    Unlike solely collecting newly released data as before,
    it constructs test samples based on identified updated real-world knowledge,
    ensuring strictly contamination-free evaluation.
    It also introduces a fully automated building workflow without human labor.
    This enables frequent and efficient benchmark updates to accommodate emerging LLMs, greatly simplifying maintenance.
    These advantages establish \benchname as an ideal testbed for contamination-free evaluation,
    providing accurate and fair assessments for LLMs.

\section*{Limitations}
    Our benchmarking framework can ensure strictly contamination-free evaluation with a fully automated building workflow,
    but we believe there are some limitations to be explored as future work:
    \begin{itemize}[leftmargin=*, itemsep=2pt]
        \item
            To implement the automated building workflow,
            we mainly consider the question-answering tasks for evaluation.
            Although we have devised different task difficulty levels,
            more diverse tasks can be explored in the future, which will more extensively evaluate LLMs in a contamination-free manner.
            We think the challenges lie in how to identify the knowledge embedded in these tasks and how to automatically construct new high-quality samples.
        \item
            Our benchmarking framework uses Wikidata and Wikipedia as data sources.
            While these sources are extensive and frequently updated by their contributor communities, they may contain incorrect information in rare cases.
            As discussed in \Cref{sec_human_verifycation}, most produced samples are correct as verified by humans.
    \end{itemize}

\section*{Ethics Statement}
    In this paper, we build and maintain our \benchname with Wikidata and Wikipedia as the data sources.
    We acknowledge that Wikidata and Wikipedia may contain inaccurate information in a few cases as they are extremely abundant and rely on human labor for maintenance.
    Besides, our work focuses on real-world commonsense knowledge by sampling relations with physical entities (See \Cref{app_workflow}). 
    This can avoid harmful information in most situations.

\bibliography{lib}

\clearpage
\appendix

\begin{table*}[!t]
    \centering
    \setlength{\tabcolsep}{3mm}
    \renewcommand{\arraystretch}{1.2}
    \resizebox{0.8\linewidth}{!}{
        \begin{tabular}{l|rrrr|rrrr}
        \toprule
        \multicolumn{1}{c|}{\multirow{2}[4]{*}{\textbf{Time period}}} & \multicolumn{4}{c|}{\textbf{Single-Hop}} & \multicolumn{4}{c}{\textbf{Multi-Hop}} \\
        \cmidrule{2-9}      & \multicolumn{1}{c}{Gold} & \multicolumn{1}{c}{$N_{\mscript{d}}$=3} & \multicolumn{1}{c}{$N_{\mscript{d}}$=5} & \multicolumn{1}{c|}{$N_{\mscript{d}}$=7} & \multicolumn{1}{c}{Gold} & \multicolumn{1}{c}{$N_{\mscript{d}}$=3} & \multicolumn{1}{c}{$N_{\mscript{d}}$=5} & \multicolumn{1}{c}{$N_{\mscript{d}}$=7} \\
        \midrule
        2022-01-01 to 2023-01-01 & 1090  & 1089  & 1088  & 1088  & 443   & 443   & 443   & 443 \\
        2023-05-01 to 2024-08-01 & 819   & 818   & 818   & 818   & 941   & 939   & 939   & 939 \\
        \bottomrule
        \end{tabular}%
    }
    \caption{
        Sample sizes in the constructed \benchname in the experiments.
    }
    \label{tab_stats_size}
\end{table*}

\begin{table*}[!t]
    \centering
    \setlength{\tabcolsep}{3mm}
    \renewcommand{\arraystretch}{1.2}
    \resizebox{0.8\linewidth}{!}{
        \begin{tabular}{l|rrrr|rrrr}
        \toprule
        \multicolumn{1}{c|}{\multirow{2}[4]{*}{\textbf{Time period}}} & \multicolumn{4}{c|}{\textbf{Single-Hop}} & \multicolumn{4}{c}{\textbf{Multi-Hop}} \\
        \cmidrule{2-9}      & \multicolumn{1}{c}{Gold} & \multicolumn{1}{c}{$N_{\mscript{d}}$=3} & \multicolumn{1}{c}{$N_{\mscript{d}}$=5} & \multicolumn{1}{c|}{$N_{\mscript{d}}$=7} & \multicolumn{1}{c}{Gold} & \multicolumn{1}{c}{$N_{\mscript{d}}$=3} & \multicolumn{1}{c}{$N_{\mscript{d}}$=5} & \multicolumn{1}{c}{$N_{\mscript{d}}$=7} \\
        \midrule
        2022-01-01 to 2023-01-01 & 5998  & 23163 & 33867 & 46033 & 24646 & 40611 & 50846 & 61761 \\
        2023-05-01 to 2024-08-01 & 7210  & 27501 & 40800 & 54451 & 25505 & 43926 & 53898 & 66957 \\
        \bottomrule
        \end{tabular}%
    }
    \caption{
        Average word counts of samples in the constructed \benchname in the experiments.
    }
    \label{tab_stats_word_count}
\end{table*}

\section{Building Workflow Details} \label{app_workflow}
    We use the Wikidata dump released on 2024-08-05 as our data source.
    To sample relations from Wikidata,
    we browse the relation list from Wikidata~\cite{wikidata2014} ~\footnote{\url{https://www.wikidata.org/wiki/Wikidata:Database_reports/List_of_properties/all}}
    and manually select common relations associated with physical entities and exclude the less meaningful ones, such as those related to virtual entries (\eg geographic coordinates and IMDB ID).
    The sampled relations mainly focus on common topics, such as sports, politics, and entertainment
    \cite{wu2020short,wu2022mitigating,wu2023infoctm,wu2024dynamic,wu2024traco,wu2024topmost,wu2024survey,wu2024fastopic}.
    We predefine the question templates for each sampled relation,
    for instance, \textit{What sports team is {} a member of?} for the relation \textit{member of sports team}.
    For more details, see the configuration files in our code.
    We follow simple-wikidata-db~\footnote{\url{https://github.com/neelguha/simple-wikidata-db}}
    and
    extract the claims of sampled relations, qualifiers, aliases, and Wikipedia titles from the dump.
    To find updated knowledge, we combine the claims and their \textit{start time} and \textit{end time} qualifiers
    and then sort them by \textit{start time}.
    We check if the object changes after the preset cutoff time and identify updated knowledge if it changes, for instance, Messi's sports team changed in \Cref{fig_workflow}.

    Given an entity, we employ the MediaWiki APIs~\footnote{\url{https://www.mediawiki.org/wiki/MediaWiki}} and use its extracted Wikipedia title to retrieve its Wikipedia article and revision history.
    Note that our workflow supports building multilingual benchmarks.
    We only need to prepare the configuration files in another language and then execute the workflow by specifying that language.
    The workflow automatically retrieves corresponding Wikipedia articles in that language.

    The statistics of produced benchmarks are reported in \Cref{tab_stats_size,tab_stats_word_count}.

\begin{table}[!t]
    \centering
    \renewcommand{\arraystretch}{1.2}
    \resizebox{\linewidth}{!}{
        \begin{tabular}{lrr}
        \toprule
        \textbf{Model} & \textbf{Release time} & \textbf{Knowledge cutoff time} \\
        \midrule
        Llama-2-7B & 2023-07 & 2022-09 \\
        Llama-2-13B & 2023-07 & 2022-09 \\
        Mistral-7B & 2023-09 & 2022* \\
        Vicuna-v1.5-7B & 2023-07 & 2022-09 \\
        Longchat-v1.5-7B & 2023-07 & 2022-09 \\
        \midrule
        Llama-3.1-8B & 2024-07 & 2023-12 \\
        Phi-3.5-mini & 2024-08 & 2023-10 \\
        Qwen-2-7B & 2024-06 & 2023* \\
        Mistral-Nemo-12B & 2024-07 & 2024-04 \\
        Gemma-2-9B & 2024-08 & 2024-06* \\
        GPT-4o-mini & 2024-07 & 2023-10 \\
        GPT-4o & 2024-07 & 2023-12 \\
        \bottomrule
        \end{tabular}%
    }
    \caption{
        Release time and knowledge cutoff time of LLMs.
        * means estimated time since some LLMs do not disclose their cutoff time.
    }
    \label{tab_LLM_date}
\end{table}

\begin{table}[!t]
    \centering
    \resizebox{\linewidth}{!}{
        \begin{tabular}{lrrrr}
        \toprule
        \multirow{2}[4]{*}{\textbf{Quality Metrics}} & \multicolumn{2}{c}{\textbf{Single-Hop Gold}} & \multicolumn{2}{c}{\textbf{Multi-Hop Gold}} \\
        \cmidrule{2-5}      & \multicolumn{1}{c}{\makecell{Raw\\ Agreement}} & \multicolumn{1}{c}{\makecell{Gwet's \\ AC1}} & \multicolumn{1}{c}{\makecell{Raw\\ Agreement}} & \multicolumn{1}{c}{\makecell{Gwet's \\ AC1}} \\
        \midrule
        Context Accuracy & 0.98  & 0.99  & 0.97  & 0.98 \\
        Answer Accuracy & 0.97  & 0.98  & 0.96  & 0.97 \\
        \bottomrule
        \end{tabular}%
    }
    \caption{Annotation agreement analysis.}
    \label{tab_annotation_agreement}
\end{table}

\section{Data Quality Verification} \label{app_data_verification}
    We recruit three annotators (graduate students) to verify the data quality of our benchmark.
    Specifically, we provide 100 samples of Single-Hop Gold and 100 samples of Multi-Hop Gold;
    then we ask annotators to determine
    (i) if the provided context can sufficiently answer the question;
    (ii) if the given answer is accurate and can appropriately address the question given the context.
    \Cref{tab_data_quality} summarizes the average accuracy scores concerning Single-Hop Gold and Multi-Hop Gold.

    \Cref{tab_annotation_agreement} reports the inter-annotator agreement analysis.
    We mainly use two agreement coefficients: Raw Agreement and Gwet's AC1.
    This is because the annotations are extremely unbalanced (most of them are positive); other coefficients like Krippendorff’s Alpha and Fleiss’ Coefficient may be inappropriate \cite{gwet2008computing}.
    It shows that the annotations maintain high agreement.

\clearpage
\section{Prompts} \label{app_prompt}

    Following \citet{bai2024longbench},
    we use the following prompts for generation and multi-choice question formats mentioned in \Cref{sec_construct_samples}.

\lstset{
    style=prompt_style,
}
\begin{lstlisting}
You are given an article and a question. Answer the question based on the given article as concisely as you can, using a single phrase or sentence if possible. Do not provide any explanation.

(*@\color{codepurple}{\textbf{Article}}@*): <context>

(*@\color{codepurple}{\textbf{Question}}@*): <question>

(*@\color{codepurple}{\textbf{Answer}}@*):
\end{lstlisting}

\lstset{
    style=prompt_style,
}
\begin{lstlisting}
You are given an article, a question, and four options. Select one option to answer the question based on the given article. Only give the option (A, B, C, or D), and do not output any other words.

(*@\color{codepurple}{\textbf{Article}}@*): <context>

(*@\color{codepurple}{\textbf{Question}}@*): <question>
<options>

(*@\color{codepurple}{\textbf{Answer}}@*):
\end{lstlisting}

\clearpage
\onecolumn

\section{Examples of \benchname} \label{app_examples}

We list some examples in the \benchname. \\

\textbf{Examples of Single-Hop Gold}

\begin{lstlisting}[language=json]
[
    {
         "question": "What sports team is Duncan Cowan Ferguson a coach of?",
         "answer": [
             "Inverness Caledonian Thistle F.C.",
             "Inverness Caledonian Thistle Football Club",
             "Inverness Caledonian Thistle FC",
             "Inverness Caledonian Thistle",
             "Inverness",
             "ICTFC",
             "Caley Thistle"
         ],
         "context": "Duncan Cowan Ferguson (born 27 December 1971) is a Scottish football coach and former player who is the manager of Scottish Championship club Inverness Caledonian Thistle..."
    },
    {
        "question": "What is the position held by Leo Docherty?",
        "answer": [
            "Minister of State for the Armed Forces"
        ],
        "context": "Leo Docherty (born 4 October 1976) is a British politician serving as Minister of State for the Armed Forces since 26 March 2024..."
    },
    {
        "question": "What sports team is Keiren Westwood a member of?",
        "answer": [
            "Queens Park Rangers F.C.",
            "Queens Park Rangers Football Club",
            "Queens Park Rangers FC",
            "Queens Park Rangers",
            "The Hoops",
            "The Rs",
            "QPRFC"
        ],
        "context": "Keiren Westwood (born 23 October 1984) is a professional footballer who plays as a goalkeeper for Queens Park Rangers. Born in England, he plays international football for the Republic of Ireland..."
    }
]
\end{lstlisting}

\clearpage
\textbf{Examples of Multi-Hop Gold}

\begin{lstlisting}[language=json]
[
    {
        "question": "What is the headquarter of the sports team that Kevin Luckassen is a member of?",
        "answer": [
            "Arad",
            "Arad, Romania"
        ],
        "context": [
            "Passage 1: Kevin Luckassen (born 27 July 1993) is a Dutch professional footballer who plays as a forward for Romanian Liga I club UTA Arad...",
            "Passage 2: Asociatia Fotbal Club UTA Arad (), commonly known as UTA Arad or simply UTA (Uzina Textila Arad (\"Textiles Factory of Arad\")), is a Romanian professional football club based in the city of Arad, Romania, which competes in the Liga I..."
        ]
    },
    {
        "question": "Who is the spouse of the officeholder of President of Finland?",
        "answer": [
            "Suzanne Innes-Stubb",
            "Suzanne Innes",
            "Suzanne Stubb",
            "Suzanne Elizabeth Innes-Stubb",
            "Suzanne Elizabeth Innes"
        ],
        "context": [
            "Passage 1: Cai-Goran Alexander Stubb (born 1 April 1968) is the 13th and current President of Finland, having won the 2024 presidential election. He previously served as Prime Minister of Finland from 2014 to 2015...",
            "Passage 2: Suzanne Elizabeth Innes-Stubb (born 25 January 1970) is a British-Finnish attorney and the wife of Alexander Stubb, President of Finland. She was the first person of overseas origin to become the spouse of the President of Finland..."
        ]
    },
    {
        "question": "What is the country of citizenship of the head coach of the sports team that Marquez Reshard Valdes-Scantling is a member of?",
        "answer": [
            "United States of America",
            "United States",
            "American",
            "Americans"
        ],
        "context": [
            "Passage1: Marquez Reshard Valdes-Scantling (born October 10, 1994) is an American football wide receiver for the Kansas City Chiefs of the National Football League (NFL). He played college football at NC State and South Florida, and was drafted by the Packers in the fifth round of the 2018 NFL Draft.",
            "Passage 2: Andrew Walter Reid (born March 19, 1958) is an American football coach who is the head coach for the Kansas City Chiefs of the National Football League (NFL). Reid was previously the head coach of the Philadelphia Eagles, a position he held from 1999 to 2012. From 2001 to 2012, he was also the Eagles' executive vice president of football operations, making him the team's general manager. He is the only NFL coach to win 100 games and appear in four consecutive conference championships with two different franchises."
        ]
    }
]
\end{lstlisting}

\end{document}